# Multi-Activation Hidden Units for Neural Networks with Random Weights

Ajay M. Patrikar
ajay.m.patrikar@gmail.com

*Abstract*—Single layer feedforward networks with random weights are successful in a variety of classification and regression problems. These networks are known for their non-iterative and fast training algorithms. A major drawback of these networks is that they require a large number of hidden units. In this paper, we propose the use of multi-activation hidden units. Such units increase the number of tunable parameters and enable formation of complex decision surfaces, without increasing the number of hidden units. We experimentally show that multi-activation hidden units can be used either to improve the classification accuracy, or to reduce computations.

## I. Introduction

Single layer feedforward networks with random weights have been studied since the early nineties [1-4] and have been successfully applied to a large number of pattern classification and regression problems in the last two decades. A survey of these networks can be found in [5]. In the literature, these networks have often been referred to as random vector functional link (RVFL) networks [6-10] or extreme learning machines (ELM) [11-12]. In this paper, we will refer to them as neural networks with random weights (NNRW). These networks are characterized by random assignment of hidden unit weights, which are not trained. Weights between the hidden layer and the output layer are analytically obtained using non-iterative training algorithms. These algorithms [5] are known to be much faster than the conventional neural networks, which depend on iterative training algorithms based on error backpropagation.

A known drawback of NNRW is the large number of hidden units required by these networks to achieve good accuracy. This can result in a longer running time during inference, which can limit their use on platforms with limited computational power such as embedded systems, Internet of Things, smartphones, drones, etc. Presently, machine learning algorithms are increasingly being adopted on such platforms, emphasizing the need for efficient machine learning models.

There have been several attempts to reduce the number of hidden units reported in the literature [11-19]. These methods often depend on incrementally adding or pruning hidden units in the network. In this paper, we take a very different approach. We attempt to reduce the number of hidden units by using multiple activations per hidden unit. A similar method based on activation ensemble has recently been investigated in the context of deep networks [28, 29]. Our proposed method is simpler and is focused on NNRW. Another method which uses multiple activations for NNRW ensemble was proposed in [30]. Our proposed method differs in that it does not use NNRW ensemble. Instead, it uses multiple activations per hidden unit in a single NNRW. Use of multiple activation functions allows for the formation of varied decision surfaces. We experimentally show that multiple activations lead to improved classification accuracy. Alternatively, this method can help reduce the number of hidden units leading to reduced computations.

The paper is organized as follows. Section II introduces NNRW with multi-activation hidden units. In Section III, experimental results are presented on a number of benchmark machine learning problems. Section IV gives a summary and conclusions.

## II. Multi-Activation Hidden Units

A single layer feedforward network is shown in Fig. 1. The hidden layer has been modified to include two or more activation functions per summation unit. There are no weights associated with the links between the summation units and the activation functions. In the case of NNRW, the weights between the input and hidden layers are chosen randomly, and only the weights between the hidden and output layers are trained. If only one activation function is used per summation unit, this model leads to traditional implementation of NNRW. In some implementations of NNRW, there are direct connections between the input and output layer (e.g. RVFL). While the proposed method is applicable to those implementations, we focus our analysis only on the architecture shown in Fig. 1.

Let $\vec{x}$ be the input feature vector. Let $d_i^n(\vec{x})$ be the output of the $n$th activation function $g^n(\ )$ of the $i$th hidden unit:

$$d_i^n(\vec{x}) = g^n(\vec{a}_i \cdot \vec{x} + b_i) \qquad (1)$$

where $\vec{a}_i$ is the random weight vector and $b_i$ is the bias term associated with the $i$th hidden unit. We construct a vector $\vec{h}(\vec{x})$ as below:

$$\vec{h}(\vec{x}) = [d_1^1(\vec{x}), \dots, d_1^{N_A}(\vec{x}), d_2^1(\vec{x}), \dots, d_2^{N_A}(\vec{x}), \dots, d_M^N(\vec{x}), \dots, d_M^{N_A}(\vec{x})]. \qquad (2)$$

The dimension of this vector is $N_A * M$, where $M$ is the number of summation units and $N_A$ is the number of activation functions per summation unit. We define the output function for each class to be

$$f^n(\vec{x}) = \vec{h}(\vec{x}) \cdot \vec{\beta}^n \quad (3)$$

where $\vec{\beta}^n = [w_1^n, w_2^n, \ldots, w_{N_A*M}^n]^T$ is a vector of the output weights for the *n*th class. Our goal is to determine the output weights $\vec{\beta}^n$ for each class.

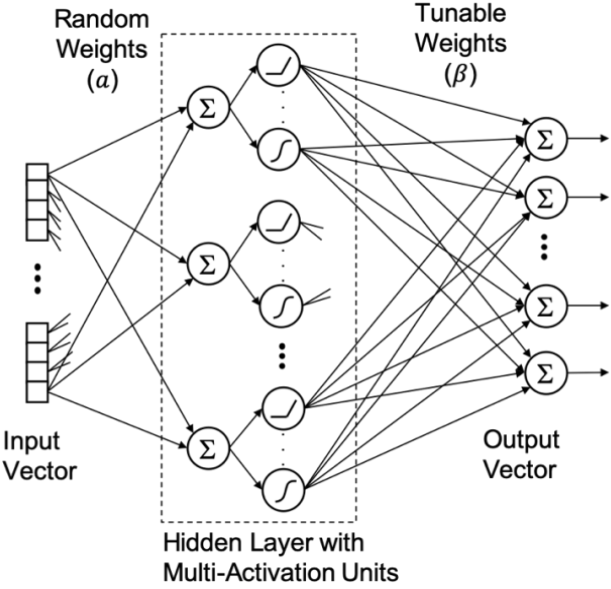

Fig. 1. Single layer feedforward network with random weights and multi-activation hidden units

Given $L$ training samples $\{(\vec{x}_i, \vec{t}_i)\}_{i=1}^L$, we seek a solution to the following learning problem:

$$\boldsymbol{H \beta = T} \quad (4)$$

where $\boldsymbol{T} = [\vec{t}_1, \ldots, \vec{t}_L]^T$ are target labels, $\boldsymbol{H} = [\vec{h}(\vec{x}_1), \ldots, \vec{h}(\vec{x}_L)]^T$ is a matrix consisting of hidden unit output vectors, and $\boldsymbol{\beta} = [\vec{\beta}^1, \ldots, \vec{\beta}^P]^T$ is the output weight matrix. There are $P$ classes. The output weights $\boldsymbol{\beta}$ can be calculated as follows:

$$\boldsymbol{\beta = H^\dagger T} \quad (5)$$

where $\boldsymbol{H}^\dagger$ is the Moore-Penrose generalized inverse of matrix $\boldsymbol{H}$. There are several methods for calculation of $\boldsymbol{H}^\dagger$. These include the orthogonal projection method, orthogonalization method, iterative method, and singular value decomposition [20-21]. Another alternative is to use ridge regression [22-23] for which a solution is given by

$$\boldsymbol{\beta = H(H^T H + \lambda I)^{-1} T} \quad (6)$$

where $\boldsymbol{I}$ is the $(N_A * M) \times (N_A * M)$ identity matrix and $\lambda$ is a tunable parameter. We use this method in our experiments with $\lambda$ set to 0.01. For a network of $N$ inputs, $M$ hidden units, $N_A$ activations per hidden unit, and $P$ outputs, the number of multiply-and-accumulate arithmetic operations during inference are approximately $(N * M + M * N_A * P)$, if the bias terms are ignored. We use this formula to compare network computations.

In NNRW, the random projection performed by the hidden layer usually does not contain any information specific to the classification or regression problem that the network is trying to solve. However, there is significant amount of computation associated with the random projection step. By sharing the hidden units between activations, we limit the number of random parameters and associated computations. At the same time, we increase the number of tunable weights $\boldsymbol{\beta}$ by introducing multiple activations in the hidden layer. Increase in tunable weights often leads to better performance, until overfitting causes the performance to deteriorate.

Using different types of activation functions has another advantage. It allows formation of more complex decision surfaces, which can enhance the classification capabilities of the network. In NNRW, the activation functions need not be differentiable; therefore, there are many nonlinear functions to choose from. The popular ones include sigmoid, tanh, Gaussian, rectified linear units (ReLU), leaky ReLU, etc. Other functions such as hardlim, sine, tribas, cubic, and signed-quadratic functions have also been used [6, 20]. An entirely new class of activation functions for deep networks have been recently investigated in [24, 29]. In the next section, we experimentally verify that factors such as more tunable parameters and more complex decision surfaces lead to either better accuracy or a smaller network.

TABLE I. TRANSFER FUNCTIONS FOR VARIOUS ACTIVATIONS

| Activation Function | Formula |
|---|---|
| Sigmoid | $f(y) = \dfrac{1}{1 + e^{-y}}$ |
| Gaussian | $y = e^{-y^2}$ |
| Leaky ReLU | $f(y) = \begin{cases} y & y > 0 \\ 0.2y & y \leq 0 \end{cases}$ |

### III. EXPERIMENTAL RESULTS

In this section, we describe experiments on three benchmark machine learning problems. In each case, a two-activation NNRW is compared with the baseline NNRW models, which make use of only one activation function per summation unit. The activation functions used in our experiments are chosen from Table I.

#### A. Results of the SatImage Problem

NNRW classifiers were trained for the Landsat satellite image (SatImage) problem from the Statlog [25] collection. This problem contains 36 attributes, six classes, 4,435 training samples, and 2,000 test samples. Twenty-five trials were conducted with different random initializations and the average classification accuracy was calculated. Fig. 2 shows the average classification accuracy as a function of the number of hidden units. The best accuracy of 90.68% was obtained using a two-activation NNRW (Sigmoid + Gaussian) with 500 hidden units. The performance drop beyond 500 units for this network was due to overfitting. The highest accuracy obtained by the single-activation NNRWs was 90.51% with 1,000 units. The top performing two-activation NNRW requires 43% fewer computations during inference than the computations required by the top performing single-activation NNRW.

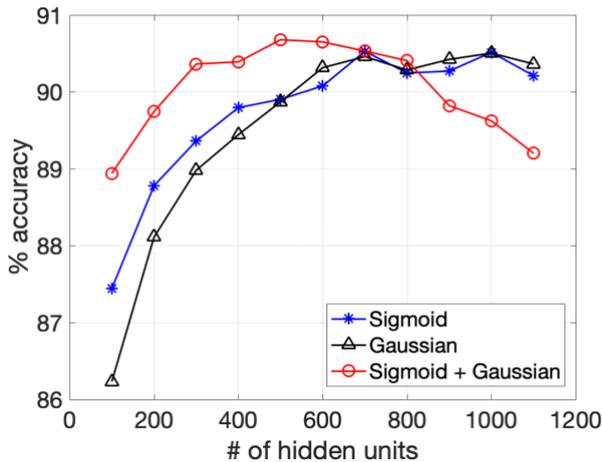

Fig. 2. Results of SatImage classification problem

### B. Results of the UCI Letter Recognition Problem

The UCI letter recognition problem [25] contains 16 attributes and 26 classes. The data consist of 20,000 samples. For each trial, the training data set and test data set are randomly generated from the overall database. 13,333 samples were used for training and 6,667 samples were used for testing. Twenty-five trials were conducted with different random initializations as well as data partitions, and the average classification accuracy was calculated. The results are shown in Fig. 3. The best accuracy obtained by the two-activation NNRW (Sigmoid + Gaussian) was 96.74% with 2,600 hidden units. It can be seen from Fig. 3 that the two-activation NNRW consistently outperforms the single-activation NNRWs. The best accuracy obtained by the single-activation NNRWs was 96.19% with 2,600 hidden units. In comparison, a two-activation NNRW with 1,400 hidden units had an accuracy of 96.22%, which is about a 14% drop in computations during inference.

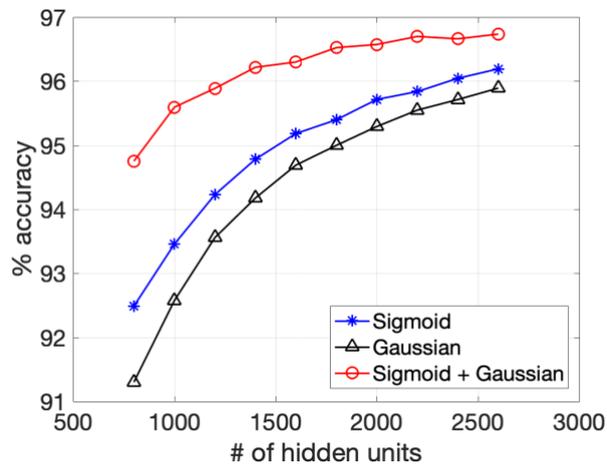

Fig. 3. Results of the UCI letter recognition problem

### C. Results of the MNIST Classification Problem

MNIST is a benchmark problem for handwritten digit recognition [26]. The problem consists of 10 classes, 60,000 training images, and 10,000 test images. The dimensionality of images is 28x28 pixels. We used the original MNIST dataset without any distortions. Thus, the dimensionality of the feature vector was 784. For this problem, we made use of shaped input weights to initialize hidden layer weights as described in [22, 27] which are known to provide better accuracy. The results averaged over twenty-five trials are shown in Fig. 4. The best accuracy obtained was 98.96% using a two-activation NNRW with 7,000 hidden units. It can be seen that the two-activation NNRW consistently outperforms single-activation NNRWs. The best accuracy obtained by the single-activation NNRWs was 98.76% with 7,000 hidden units. In comparison, a two-activation NNRW had an accuracy of 98.71% with 3,000 hidden units, which is about a 56% drop in computations during inference.

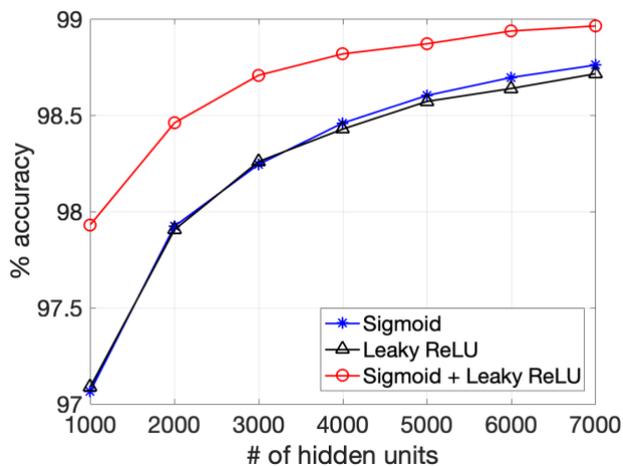

Fig. 4. Results of the MNIST classification problem

## IV. CONCLUSION

We have experimentally shown that using multi-activation hidden units in NNRW results in overall superior performance. The proposed method can be used either to improve accuracy or to reduce computations. While our experiments are limited to two activation functions, it is certainly possible to use more than two. The activation functions used in NNRW need not be differentiable. Therefore, there are many nonlinear functions to choose from. Further research is needed to determine which activation functions are complementary to each other.